\documentclass[10pt,twocolumn,letterpaper]{article}

\usepackage{cvpr}              

\usepackage{graphicx}
\usepackage{amsmath}
\usepackage{amssymb}
\usepackage{booktabs}
\usepackage{float}

\usepackage[pagebackref,breaklinks,colorlinks]{hyperref}

\usepackage[capitalize]{cleveref}
\crefname{section}{Sec.}{Secs.}
\Crefname{section}{Section}{Sections}
\Crefname{table}{Table}{Tables}
\crefname{table}{Tab.}{Tabs.}

\def\cvprPaperID{*****} 
\def\confName{CVPR}
\def\confYear{2022}

\usepackage[utf8]{inputenc} 
\usepackage[T1]{fontenc}    
\usepackage{url}            
\usepackage{booktabs}       
\usepackage{amsfonts}       
\usepackage{nicefrac}       
\usepackage{microtype}      
\usepackage{xcolor}         
\usepackage{multirow} 
\usepackage{enumitem}
\usepackage{adjustbox}
\usepackage{subcaption}
\usepackage{caption}
\usepackage{pifont}
\usepackage{algorithm}
\usepackage{algorithmic}
\usepackage{wrapfig}
\usepackage{sidecap}
\usepackage{amssymb}
\usepackage{cuted}
\definecolor{ForestGreen}{RGB}{34,139,34}
\definecolor{Cerulean}{RGB}{42,82,190}
\definecolor{CornflowerBlue}{RGB}{100,149,237}
\definecolor{Turquoise}{RGB}{48,213,200}
\definecolor{ProcessBlue}{RGB}{0,136,208}
\usepackage{tabularx}
\definecolor{lightgray}{gray}{0.9}
\definecolor{lightblue}{rgb}{0.93,0.95,1.0}
\definecolor{darkgreen}{rgb}{0.0,0.6,0.0}
\definecolor{darkblue}{rgb}{0.0,0.0,0.5}
\definecolor{pinegreen}{rgb}{0.0, 0.47, 0.44}
\definecolor{deepmagenta}{rgb}{0.8, 0.0, 0.8}
\definecolor{amber}{rgb}{1.0, 0.49, 0.0}

\definecolor{patchTokensBlue}{rgb}{0.56, 0.67, 0.86}
\definecolor{objTokensPurple}{rgb}{0.44, 0.19, 0.63}
\definecolor{objPromptsPurple}{rgb}{0.77, 0.54, 0.86}

\newcommand{\cmark}{\textcolor{darkgreen}{\ding{51}}}
\newcommand{\xmark}{\textcolor{red}{\ding{55}}}

\newcommand\drop[1]{}

\newcommand{\reals}{\mathbb{R}}

\newcommand{\bb}{\boldsymbol{b}}

\newcommand{\ignorebig}[1]{}

\newcommand{\minisection}[1]{\noindent{\textbf{#1}.}}
\newcommand{\tabref}[1]{Table~\ref{#1}}
\newcommand{\eqrref}[1]{Equation~\ref{#1}}

\newcommand{\tablestyle}[2]{\setlength{\tabcolsep}{#1}\renewcommand{\arraystretch}{#2}\centering\footnotesize}
\newlength\savewidth

\newcommand{\methodwithoutboxes}{} 

\newcommand{\graph}{HAOG}
\newcommand{\model}{SViT}
\newcommand{\approach}{SViT}

\newcommand{\loss}{frame-clip Consistency}

\definecolor{citecolor}{RGB}{34,139,34}
\definecolor{lightred}{RGB}{241,140,142}
\definecolor{amber(sae/ece)}{rgb}{1.0, 0.49, 0.0}
\definecolor{battleshipgrey}{rgb}{0.52, 0.52, 0.51}
\definecolor{cadmiumorange}{rgb}{0.93, 0.53, 0.18}
\definecolor{darkorchid}{rgb}{0.6, 0.2, 0.8}

\newcommand\patchTokensBlue[1]{\textcolor{patchTokensBlue}{\textbf{#1}}}
\newcommand\objTokensPurple[1]{\textcolor{objTokensPurple}{\textbf{#1}}}
\newcommand\objPromptsPurple[1]{\textcolor{objPromptsPurple}{\textbf{#1}}}

\begin{document}

\title{Structured Video Tokens @ Ego4D PNR Temporal Localization Challenge 2022}

\author{
Elad Ben-Avraham$^{1}$
\\ {\tt\small eladba4@gmail.com}
\and
Roei Herzig$^{1,3}$
\\ {\tt\small roeiherz@gmail.com}
\and
Karttikeya Mangalam$^{2}$
\\ {\tt\small mangalam@cs.berkeley.edu}
\and
Amir Bar$^{1}$
\\ {\tt\small amirb4r@gmail.com}
\and
Anna Rohrbach$^{2}$
\\ {\tt\small anna.rohrbach@berkeley.edu}
\and
Leonid Karlinsky$^{4}$
\\ {\tt\small leonidka@ibm.com}
\and
Trevor Darrell$^{2}$
\\ {\tt\small trevordarrell@berkeley.edu}
\and
Amir Globerson$^{1,5}$
\\ {\tt\small gamir@tauex.tau.ac.il}
\and
\\
$^1$Tel Aviv University, $^2$UC Berkeley, $^3$IBM Research, $^4$MIT-IBM Lab, $^5$Google Research}

\maketitle

\begin{abstract}
This technical report describes the SViT approach for the Ego4D Point of No Return (PNR) Temporal Localization Challenge. We propose a learning framework StructureViT (SViT for short), which demonstrates how utilizing the structure of a small number of images only available during training can improve a video model. SViT relies on two key insights. First, as both images and videos contain structured information, we enrich a transformer model with a set of \emph{object tokens} that can be used across images and videos. Second, the scene representations of individual frames in video should ``align'' with those of still images. This is achieved via a ``Frame-Clip Consistency'' loss, which ensures the flow of structured information between images and videos. SViT obtains strong performance on the challenge test set with 0.656 absolute temporal localization error.

\end{abstract}

\section{Introduction}
\label{sec:intro}

Semantic understanding of videos is a key challenge for machine vision and artificial intelligence. It is intuitive that video models should benefit from incorporating scene structure, e.g., the objects that appear in a video, their attributes, and the way they interact. In this work, we propose an approach, which offers an effective mechanism to leverage image structure to improve video transformers.

A natural first step towards image-video knowledge sharing is to use the same transformer model to process both. However, this still leaves two key questions: how to model structured information, and how to account for domain mismatch between images and videos. Towards this end, we introduce two key concepts: (i) A set of transformer ``object-tokens'' -- additional tokens initialized from learned embeddings (we refer to these embeddings as ``object prompts'') that are meant to capture object-centric information in both still image and video. These tokens can be supervised based on the scene structure of images, and that information will also be propagated to videos. To formalize the image structure, we proposed the HAnd-Object Graph ({\graph}), a simple representation of the interactions between hands and objects in the scene. (ii) A novel ``Frame-Clip Consistency loss'' that ensures consistency between the ``object-tokens'' when they are part of a video vs. a still image. We name our proposed approach StructureViT (\emph{Bringing Scene \textbf{Structure} from Images to \textbf{Vi}deo via Frame-Clip Consistency of Object \textbf{T}okens}, or {\approach} for short).

\begin{figure}[ht]
    \centering
    \includegraphics[width=.7\linewidth]{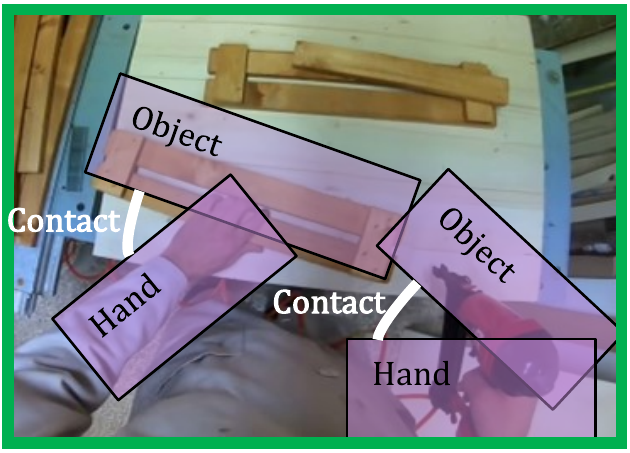}
    \captionof{figure}{Visualization of {\graph} annotation on a still image.}
    \label{fig:haog}
\end{figure}


\begin{figure*}
    \includegraphics[width=1.0\linewidth]{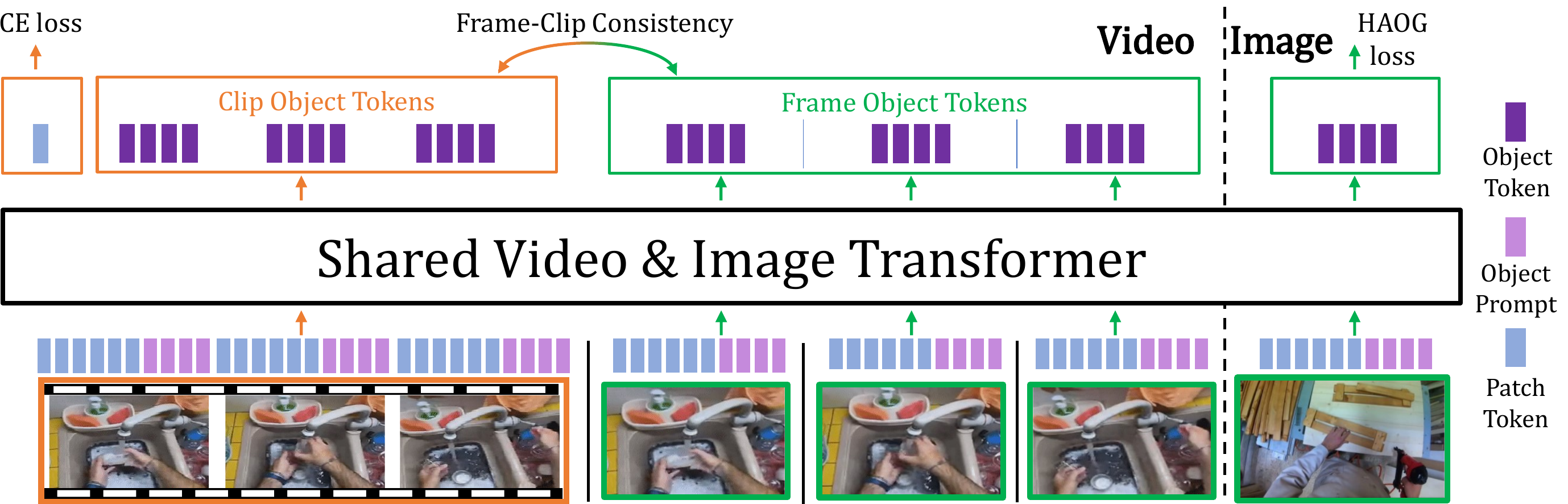}
    \captionof{figure}{Our shared video \& image transformer model processes two different types of tokens: standard patch tokens from the images and videos (\patchTokensBlue{blue}) and the object prompts (\objPromptsPurple{purple}), that are transformed into object tokens (\objTokensPurple{purple}) in the output. During training, the object tokens (\objTokensPurple{purple}) are trained to predict the {\graph} for still images. For video frames that have no {\graph} annotation, we use our ``Frame-Clip'' loss to ensure consistency between the ``frame object tokens'' (resulting from processing the frames separately) and the ``clip object tokens'' (resulting from processing the frame as part of the video). Last, the final video downstream task prediction results from applying a video downstream task head on the average of the patch tokens in the transformer output (after they have interacted with the clip object tokens.
    }
    
    \label{fig:method}
\end{figure*}

\section{The {\approach} Approach}
\label{sec:method}

Our approach {\approach} learns a structured shared representation that arises from the interaction between the two modalities of images and videos. We consider the setting where the main goal is to learn video understanding downstream tasks while leveraging structured image data. In training, we assume that we have access to task-specific video labels and structured scene annotations for images, \textit{Hand-Object Graph}s. The inference is performed only for the downstream video task without explicitly using the structured image annotations.

\subsection{The Hand-Object Graph}
\label{sec:method:haog}

We propose to use a graph-structure we call \textit{Hand-Object Graph} ({\graph}) (see \cref{fig:haog}). The nodes of the {\graph} represent two hands and two objects with their locations, whereas the edges correspond to physical properties such as contact/no contact. Formally, an {\graph} is a tuple $(O, E)$ defined as follows: \textbf{Nodes $\mathbf{O}$:} A set $O$ of $n$ objects. We assume that each image comes with the bounding boxes describing the locations of the two hands, and the objects they interact with. The bounding boxes for the left and right hand are denoted as $\bb_1,\bb_2\in\reals^4$, respectively, while for the two corresponding objects interacting with the hands are denoted as $\bb_3,\bb_4$. We also assume access to four binary variables $e_1,\ldots,e_4\in\{0,1\}$ that indicate whether the corresponding hands or objects in fact exist in the image. \textbf{Edges $\mathbf{E}$:} The edges are represented as labeled directed edges between hand and object nodes. Each edge is characterized by a physical property of ``contact'' or ``no-contact''. We denote two binary variables $c_1, c_2\in\{0,1\}$ that specify if each of the hands is in direct contact with the corresponding object.

\subsection{The {\model} Transformer Model}
\label{sec:method:ss2v}


We next present the {\model} model: a video transformer that employs \emph{object tokens} as a means to capture structure in both images and videos. Our method is schematically illustrated in~\cref{fig:method}. We begin by reviewing the video transformer architecture, which our model extends. Next, we explain how we process both images and videos. Finally, we describe object tokens and how they are used in videos and still images.

\minisection{Shared Video \& Image Transformer} Video transformers~\cite{arnab2021vivit,gberta_2021_ICML,mvit2021,herzig2022orvit} extend the Vision Transformer model to the temporal domain. Similar to vision transformers, the input is first ``patchified'' but with temporally extended 3-D convolution (instead of 2-D) producing a down-sampled tensor $X$ of size $T \times H \times W \times d$. We refer to this as ``patch tokens''. Then, spatio-temporal position embeddings are added for providing location and time information. Next, multiple stacked self-attention blocks are applied repeatedly on the down-sampled tensor $X$ to produce the final vector representation using mean pooling over the patch tokens.

In our approach, we want to be able to process batches of images or videos. A key desideratum in this context is to be able to input both videos and still-images into the same model. Towards this end, our first change is to rely only on 2-D convolutions in the first transformer step instead of 3-D. This allows single images to be used as inputs without padding, and does not hurt performance in practice. 

\minisection{Object tokens} Our goal is to learn a shared structured representation between the video and the image domains. We do this by adding transformer tokens to represent objects. These tokens are functionally similar to the patch tokens, with two exceptions. First, they are initialized with additional trained embeddings, which we referred to as ``object prompts''. We consider the ``object prompts'' as a bridge between the images and the videos since they are being used in prediction tasks across both modalities. Second, they are used to predict the structured representations when those are available (here, we use the Hand-Object Graphs structure for still images). An object token is used to predict properties of the corresponding node in the {\graph}, and the concatenation of two object tokens is used to predict edge properties.

Next, we describe the model more formally. Let $n$ be the maximum number of modeled objects per frame (or still image). We have $n$ tokens for each frame, and thus a total of $T\times n$ tokens. The token for object $i$ at frame $t$ is initialized with the vector $\boldsymbol{o}_i + \boldsymbol{r}_t$ where $\boldsymbol{o}_i\in\reals^d$ is a learned object prompt and $\boldsymbol{r}_i\in\reals^d$ is a learned temporal positional embedding (the same one used for initializing the patch tokens). With these new tokens, we  have $T \times H\times W + T \times n$ tokens (i.e, vectors in $\reals^d$), and these together will go through the standard self-attention layers.

We denote the operator that outputs the final representation of the object tokens by $F_O$, where for a video $V$ we have $F_O(V) \in \reals ^ {T \times n \times d}$ and for an image $I$ we have $F_O(I) \in \reals ^ {n \times d}$. The $F_O(I)$ representation is used to predict structured representations for single images. We also include a loss that makes single frame representations align with those of clip representations (see \cref{method:loss}). The operator that outputs the final representation vector used for the video downstream task is denoted by $F_{CLS}$, where for video $V$ we have $F_{CLS}(V) \in \reals ^{d}$.

Finally, our method can be used on top of the most common video transformers, we use the MViTv2~\cite{li2021improvedmvit} model because it performs well empirically.

\subsection{Losses and Training}
\label{method:loss}
During training we have a set of images annotated with {\graph}s and videos with downstream task labels. We use these as inputs to our model and optimize the losses described below.
\minisection{Loss} The loss is comprised of three different terms. (i) \textbf{Video Loss}, binary cross entropy loss for frames scores. (ii) \textbf{{\graph} Loss}, the transformer outputs a set of $n$ object tokens for each image, we predict structured representations from the tokens. Recall our notation from \cref{sec:method:haog}. We let the transformer use $n=4$ object tokens which correspond to the two hands (bounding boxes $\bb_1,\bb_2$) and two manipulated objects (bounding boxes $\bb_3,\bb_4$). The $j^{th}$ token is used to predict a corresponding bounding box $b_j$, as well as the existence variable $e_j$. Furthermore, we predict the contact variable from the concatenation of the corresponding hand and object tokens. Thus, the {\graph} loss include detection loss on the bounding boxes prediction (GIoU and $L_{1}$ loss), binary cross entropy for the existence variable, and cross entropy for the contact variable.

(iii) \textbf{Frame-Clip Consistency Loss}, since we have different losses for images and video, the model could find a way to minimize the image loss in a way that only applies to the images and does not transfer to video. To avoid this, we need to make sure that the video representation contains the same type of information as the still images, and in particular the structured information carried by the object tokens. To achieve this, we explicitly enforce the object tokens to be consistent across still images (frames) and videos (clips) using a {\loss} loss. To calculate the loss, we process each video twice: once as a clip, and once as a batch of $T$ frames. Then we minimize the $L_{1}$ loss between the object tokens between the clip object tokens and the frames object tokens, after applying FC layer on the clip object tokens (following the intuition that each clip object token should contain the information from its corresponding frame object token and potentially more).

\minisection{Overall loss} Each of the three terms in the loss is multiplied by a hyper-parameter ($\lambda_{Con}$, $\lambda_{{\graph}}$, $\lambda_{Vid}$), and the total loss is the weighted combination of the three terms:
\begin{equation}
\mathcal{L}_{\text{Total}} :=  \lambda_{Con} \mathcal{L}_{Con} + \lambda_{{\graph}} \mathcal{L}_{{\graph}} +  \lambda_{Vid} \mathcal{L}_{Vid}
\label{eq:loss_all}
\end{equation}

\begin{table*}[tb!]
	\centering
	\tablestyle{10.5pt}{1.15}
	\begin{tabular}{l  c c c c c c c c }
            \toprule
             Model & Pretraining & Finetuning  & Object & Frame-Clip      & Num. & Rand & Temporal     & Test Temporal  \\
                  & Aux. Images & Aux. Images & Tokens  & Const. Loss     & Views   & Aug. & loc. error   & loc. error \\
            \midrule
            {Bi-LSTM} \methodwithoutboxes & - & - & \xmark & \xmark & 1$\times$1 & \xmark & 0.790 &  0.759 \\
            {BMN}\methodwithoutboxes    & - & - & \xmark & \xmark & 1$\times$1 & \xmark & 0.780 & 0.805 \\
            {I3D ResNet-50}\methodwithoutboxes   & - & - & \xmark & \xmark & 1$\times$1 & \xmark & 0.739 & 0.755 \\
            {MViT-v2}\methodwithoutboxes   & - & - & \xmark & \xmark & 1$\times$1 & \xmark & 0.702 & \\
            \midrule
            	
            {MViTv2 MT} & Ego4D, 100DOH & Ego4D, 100DOH & \xmark & \xmark & 1$\times$1 & \xmark & 0.678 & \\ 
            \midrule

            {\model} & Ego4D, 100DOH & - & \cmark & \cmark & 1$\times$1 & \xmark & 0.642 &  \\ 
            {\model} & 100DOH & 100DOH & \cmark & \cmark & 1$\times$1 & \xmark & 0.641 & \\
            {{\model}} & Ego4D, 100DOH & Ego4D & \cmark & \cmark & 1$\times$1 & \xmark & 0.642 & 0.656 \\ 
            \midrule
            {\model} & Ego4D, 100DOH & Ego4D & \cmark & \cmark & 1$\times$1 & \cmark & 0.640 &  \\ 
            {\model} & Ego4D, 100DOH & Ego4D & \cmark & \cmark & 1$\times$3 & \cmark & 0.640 & 0.661 \\ 
            {\model} & Ego4D, 100DOH & Ego4D & \cmark & \cmark & 3$\times$1 & \cmark & 0.642 & 0.660  \\ 
            {\model} & Ego4D, 100DOH & Ego4D & \cmark & \cmark & 3$\times$3 & \cmark & 0.641 &  \\ 
            \bottomrule
	\end{tabular}
    \caption{\textbf{Results on Ego4D PNR Temporal Localization.} Num. Views refers to the number of different temporal and spatial views during inference (\textit{Temporal}$\times$\textit{Spatial}).}
	\label{tab:ego4d}
	\vspace{-1.0em}
\end{table*}

            	


\section{Experiment}
\label{sec:expr}

\subsection{Datasets}
\label{sec:expr:datasets}
For ``auxiliary'' image datasets we use the Ego4D, and the 100 Days of Hands (100DOH)~\cite{100doh} datasets. We collected 79,921 annotated images from 100DOH, and 57,213 annotated images from Ego4D. The image annotations are based on frames selected from videos, but these are sparsely selected from the videos, so it is natural treated them as annotated still images.

\subsection{Object State Change Temporal Localization and Classification}
\label{sec:expr:ego4d}
Hands and objects are key elements in human activity. Two tasks related to hand-object interaction have recently been introduced in the Ego4D~\cite{Ego4D2021} dataset. The first is the ``Point of No Return Temporal Localization'' (or PNR temporal localization), which is defined as finding key frames in a video clip that indicate a change in object state (we refer to this key frame as PNR frame). The second is ``Object State Change Classification'' (or OSC classification), which indicates whether an object state has changed or not. In this report of focus on the PNR temporal localization task.  We use the absolute error (in seconds) between the PNR frame prediction and the actual PNR frame for evaluation.

\tabref{tab:ego4d} reports results on the PNR temporal localization task on the Ego4D dataset. We include additional baseline named ``MViTv2 Multi-Task'' (MViv2 MT), this baseline is simple multi-tasking of the PNR temporal localization task and {\graph} prediction task. We observe that {\model} improves the MViT-v2 result by $0.06$. Additionally, we experiment in using different auxiliary image datasets during the pretraining and finetuning. We also experiment with rand augment~\cite{NEURIPS2020Randaugment} and multiple spatial and temporal views during inference. We observed that applying rand augment during training improves validation error, but using multiple views (both temporal and spatial) does not.

\subsection{Implementation Details}
\label{sec:expr:impl}

Our training recipes and code are based on the MViTv2 model, and were taken from \href{https://github.com/facebookresearch/mvit}{here}. We pretrain the {\model} model on the K400~\cite{kay2017kinetics} \textit{video} dataset with our \textit{auxiliary image} datasets. Then, we finetune on the target video understanding task jointly with the \textit{auxiliary image} datasets and the {\model} loss. Each training batch contains 64 images and 64 videos in order to minimize the overall loss in~\eqrref{eq:loss_all}. 

\minisection{Optimization details} We train using $16$ frames with sample rate $4$ and batch-size $128$ (comprising $64$ videos and $64$ images) on $8$ RTX 3090 GPUs. We train our network for $10$ epochs with Adam optimizer~\cite{kingma2014adam} with a momentum of $9e-1$ and Gamma $1e-1$. Following~\cite{li2021improvedmvit}, we use $lr = 1e-5$ with half-period cosine decay. Additionally, we used Automatic Mixed Precision, which is implemented by PyTorch. 

\minisection{Training details} We use a standard crop size of $224$, and we jitter the scales from $256$ to $320$.  Additionally, we set $\lambda_{Con} = 10, \lambda_{{\graph}} = 5, \lambda_{Vid} = 1$. In order to predict the PNR frame, we predict score for each frame. Then, we consider the frame with the highest score as the prediction. As stated in \cref{method:loss}, we use binary cross entropy during training .We train the PNR temporal localization task jointly with the OSC classification task. In this technical report, we focus on the PNR temporal localization results. We also experiment with rand augment~\cite{NEURIPS2020Randaugment} during training, with magnitude set to $9$. During training, frames are sampled by selecting a random first frame and last frame, and then uniformly sampling frames between them.

\minisection{Inference details} We follow the official evaluation, available \href{https://github.com/EGO4D/hands-and-objects}{here}. Additionally, we experiment with multi temporal and spatial views during the inference. For multiple spatial views during inference, we follow \cite{li2021improvedmvit}. For multiple temporal views, instead of simple uniform sampling during inference, we include a different starting offset for each view. The aggregation of the different views is performed by choosing the frame with the maximum score.


\subsection{Negative \& Positive Examples Analysis}
\href{https://eladb3.github.io/SViT/pnr_clips/positive/}{Here}, we provide examples where {\model} successfully predicts the PNR frame. We can see that the model successfully predicts the correct PNR moments, even when there are several candidates. As an example, in the second video clip, picking up the stone may seem like a candidate for PNR. However, as we can see, the model predicted the correct PNR moment (which occurs during the process of filling the box with sand).
\href{https://eladb3.github.io/SViT/pnr_clips/negative/}{Here}, we provide examples where {\model} fails to predict the PNR frame. We can see that the {\model} predictions predict the PNR too early, when a preliminary action occurs. For example, in the second video clip, the model predicts that the PNR moment occurs when a knife is brought close to a carrot (to determine the exact location of the cut) rather than at the moment of cutting.


\vspace{-10pt}
\section{Conclusion}
\vspace{-5pt}

Video understanding is a key element of human visual perception, but modeling remains a challenge for machine vision. In this work, we demonstrated the importance of learning from a scene structure from a small set of images to facilitate video learning within or outside the domain of interest. According to our empirical study, our {\model} approach improves performance on the PNR temporal localization task. Note, that we did not prioritize making the {\graph} annotations richer, i.e., it may be possible to add other physical properties to improve structure modeling. We leave for future work how different physical properties can be more effectively utilized in more complex structures.

\subsubsection*{Acknowledgements}
This project has received funding from the European Research Council (ERC) under the European Unions Horizon 2020 research and innovation programme (grant ERC HOLI 819080). Prof. Darrell’s group was supported in part by DoD including DARPA's XAI, and LwLL programs, as well as BAIR's industrial alliance programs.

{\small
\bibliographystyle{ieee_fullname}
\bibliography{egbib}
}

\end{document}